\documentclass[letterpaper,twocolumn,fleqn]{article} 

\usepackage{ist}
\usepackage{graphics} % for pdf, bitmapped graphics files
\usepackage{epsfig} % for postscript graphics files
\usepackage{mathptmx} % assumes new font selection scheme installed
\usepackage{times} % assumes new font selection scheme installed
\usepackage{amsmath} % assumes amsmath package installed
\usepackage{amssymb}  % assumes amsmath package installed
\usepackage{fixltx2e}
\usepackage{caption}
\usepackage[colorlinks]{hyperref}
\usepackage{tikz}
\usetikzlibrary{shapes,shapes.multipart,positioning}
\usetikzlibrary{arrows,arrows.meta,graphs}
\usepackage{scrlayer-scrpage}  % header and footer for KOMA-Script

\clearscrheadfoot                 % deletes header/footer
\rehead[]{This is my name} 

\title{Deep Reinforcement Learning framework for Autonomous Driving}

\author{ Ahmad El Sallab$^{1}$, Mohammed Abdou$^{1}$, Etienne Perot$^{2}$ and Senthil Yogamani$^{3}$ \\
$^{1}$ Valeo Egypt, Cairo \\
$^{2}$ Valeo Bobigny, France  \\
$^{3}$ Valeo Vision Systems, Ireland 
}

\begin{document} 

\maketitle 
%\nocite{*}

\thispagestyle{empty} % prevents the first page to be numbered

%%%%%%%%%%%%%%%%%%%%%%%%%%%%%%%%%%
% Abstract
%%%%%%%%%%%%%%%%%%%%%%%%%%%%%%%%%%

\begin{abstract}
Reinforcement learning is considered to be a strong AI paradigm which can be used to teach machines through interaction with the environment and learning from their mistakes. Despite its perceived utility, it has not yet been successfully applied in automotive applications. Motivated by the successful demonstrations of learning of Atari games and Go by Google DeepMind, we propose a framework for autonomous driving using deep reinforcement learning. This is of particular relevance as it is difficult to pose autonomous driving as a supervised learning problem due to strong interactions with the environment including other vehicles, pedestrians and roadworks. As it is a relatively new area of research for autonomous driving, we provide a short overview of deep reinforcement learning and then describe our proposed framework. It incorporates Recurrent Neural Networks for information integration, enabling the car to handle partially observable scenarios. It also integrates the recent work on attention models to focus on relevant information, thereby reducing the computational complexity for deployment on embedded hardware. The framework was tested in an open source 3D car racing simulator called TORCS. Our simulation results demonstrate learning of autonomous maneuvering in a scenario of complex road curvatures and simple interaction of other vehicles.
\end{abstract}

%%%%%%%%%%%%%%%%%%%%%%%%%%%%%%%%%%%%
% MAIN
%%%%%%%%%%%%%%%%%%%%%%%%%%%%%%%%%%%%

%%%%%%%%%%%%%%%%%%%%%%%%%%%%%%%%%%%%%%%%%%%%%%%%%%%%%%%%%%%%%%%%%%%%%%%%%%%%%%%%
\section{INTRODUCTION}

A robot car that drives autonomously is a long-standing goal of Artificial Intelligence. Driving a vehicle is a task that requires high level of skill, attention and experience from a human driver. Although computers are more capable of sustained attention and focus than humans, fully autonomous driving requires a level of intelligence that surpasses that achieved so far by AI agents.

\begin{figure}[!hb]
	\centering
	\includegraphics[width=0.8\columnwidth]{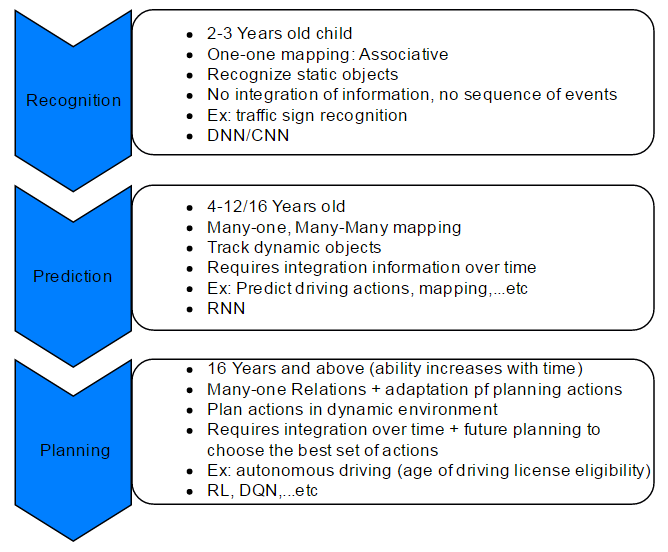}
	\caption{High level autonomous driving tasks}
	\label{fig:AItasks}
\end{figure}

The tasks involved in creating an autonomous driving agent can be divided into 3 categories, as shown in Figure \ref{fig:AItasks}:

1)	Recognition: Identifying components of the surrounding environment. Examples of this are pedestrian detection, traffic sign recognition, etc. Although far from trivial, recognition is a relatively easy task nowadays thanks to advances in Deep Learning (DL) algorithms, which have reached human level recognition or above at several object detection and classification problems \cite{c12} \cite{c3}. Deep learning models are able to learn complex feature representations from raw input data, omitting the need for hand-crafted features \cite{c21} \cite{c3} \cite{c11}. In this regard, Convolutional Neural Networks (CNNs) are probably the most successful deep learning model, and have formed the basis of every winning entry on the ImageNet challenge since AlexNet \cite{c12}. This success has been replicated in lane \& vehicle detection for autonomous driving \cite{c9}.

2)  Prediction: It is not enough for an autonomous driving agent to recognize its environment; it must also be able to build internal models that predict the future states of the environment. Examples of this class of problem include building a map of the environment or tracking an object. To be able to predict the future, it is important to integrate past information. As such, Recurrent Neural Networks (RNNs) are essential to this class of problem. Long-Short Term Memory (LSTM) networks \cite{c8} are one such category of RNN that have been used in end-to-end scene labeling systems \cite{c20}.  More recently, RNNs have also been used to improve object tracking performance in the DeepTracking model \cite{c19}.

3)  Planning: The generation of an efficient model that incorporates recognition and prediction to plan the future sequence of driving actions that will enable the vehicle to navigate successfully. Planning is the hardest task of the three. The difficulty lies in integrating the ability of the model to understand the environment  (recognition) and its dynamics (prediction) in a way that enables it to plan the future actions such that it avoids unwanted situations (penalties) and drives safely to its destination (rewards). 

The Reinforcement Learning (RL) framework \cite{c24} \cite{c29} has been used for a long time in control tasks. The mixture of RL with DL was pointed out to be one of the most promising approaches to achieve human-level control in \cite{c13}. In \cite{c17} and \cite{c16} this human-level control was demonstrated on Atari games using the Deep Q Networks (DQN) model, in which RL is responsible for the planning part while DL is responsible for the representation learning part. Later, RNNs were integrated in the mixture to account for partial observable scenarios \cite{c7}.

Autonomous driving requires the integration of information from multiple sensors. Some of them are low dimensional, like LIDAR, while others are high dimensional, like cameras. It is noteworthy in this particular example, however, that although raw camera images are high dimensional, the useful information needed to achieve the autonomous driving task is of much lower dimension. 

For example, the important parts of the scene that affect driving decisions are limited to the moving vehicle, free space on the road ahead, the position of kerbs, etc. Even the fine details of vehicles are not important, as only their spatial location is truly necessary for the problem. Hence the memory bandwidth for relevant information is much lower. If this relevant information can be extracted, while the other non-relevant parts are filtered out, it would improve both the accuracy and efficiency of autonomous driving systems. Moreover, this would reduce the computation and memory requirements of the system, which are critical constraints on embedded systems that will contain the autonomous driving control unit.

Attention models are a natural fit for such an information filtering process. Recently, these models were successfully deployed for image recognition in \cite{c32} and \cite{c15}, wherein RL was mixed with RNNs to obtain the parts of the image to attend to. Such models are easily extended and integrated to the DQN \cite{c16} and Deep Recurrent Q Networks (DRQN) \cite{c7} models. This integration was performed in \cite{c23}. The success of attention models drives us to propose them for the extraction of low level information from the raw sensory information to perform autonomous driving.

In this paper, we propose a framework for an end-end autonomous driving model that takes in raw sensor inputs and outputs driving actions. The model is able to handle partially observable scenarios. Moreover, we propose to integrate the recent advances in attention models in order to extract only relevant information from the received sensors data, thereby making it suitable for real-time embedded systems. The main contributions of this paper: 1) presenting a survey of the recent advances of deep reinforcement learning and 2) introducing a framework for end-end autonomous driving using deep reinforcement learning to the automotive community.
The rest of the paper is divided into two parts. The first part provides a survey of deep reinforcement learning algorithms, starting with the traditional MDP framework and Q-learning, followed by the the DQN, DRQN and Deep Attention Recurrent Q Networks (DARQN). The second part of the paper describes the proposed framework that integrates the recent advances in deep reinforcement learning. Finally, we conclude and suggest directions for future work.

\section{REVIEW OF REINFORCEMENT LEARNING} 

For a comprehensive overview of reinforcement learning, please refer to the second edition of Rich Sutton's textbook \cite{c25}. We provide a short review of important topics in this section. The Reinforcement Learning framework was formulated in \cite{c24} as a model to provide the best policy an agent can follow (best action to take in a given state), such that the total accumulated rewards are maximized when the agent follows that policy from the current and until a terminal state is reached. 

\subsection{Motivation for RL Paradigm}

Driving is a multi-agent interaction problem. As a human driver, it is much easier to keep within a lane without any interaction with other cars than to change lanes in heavy traffic. The latter is more difficult because of the inherent uncertainty in behavior of other drivers. The number of interacting vehicles, their geometric configuration and the behavior of the drivers could have large variability and it is challenging to design a supervised learning dataset with exhaustive coverage of all scenarios. Human drivers employ some sort of online reinforcement learning to understand the behavior of other drivers such as whether they are defensive or aggressive, experienced or in-experienced, etc. This is particularly useful in scenarios which need negotiation, namely entering a roundabout, navigating junctions without traffic lights, lane changes during heavy traffic, etc. The main challenge in autonomous driving is to deal with corner cases which are unexpected even for a human driver, like recovering from being lost in an unknown area without GPS or dealing with disaster situations like flooding or appearance of a sinkhole on the ground. The RL paradigm models uncharted territory and learns from its own experience by taking actions. Additionally, RL may be able to handle non-differentiable cost functions which can create challenges for supervised learning problems. 

\begin{figure}[htbp]
    \centering
    \includegraphics[width=1\columnwidth]{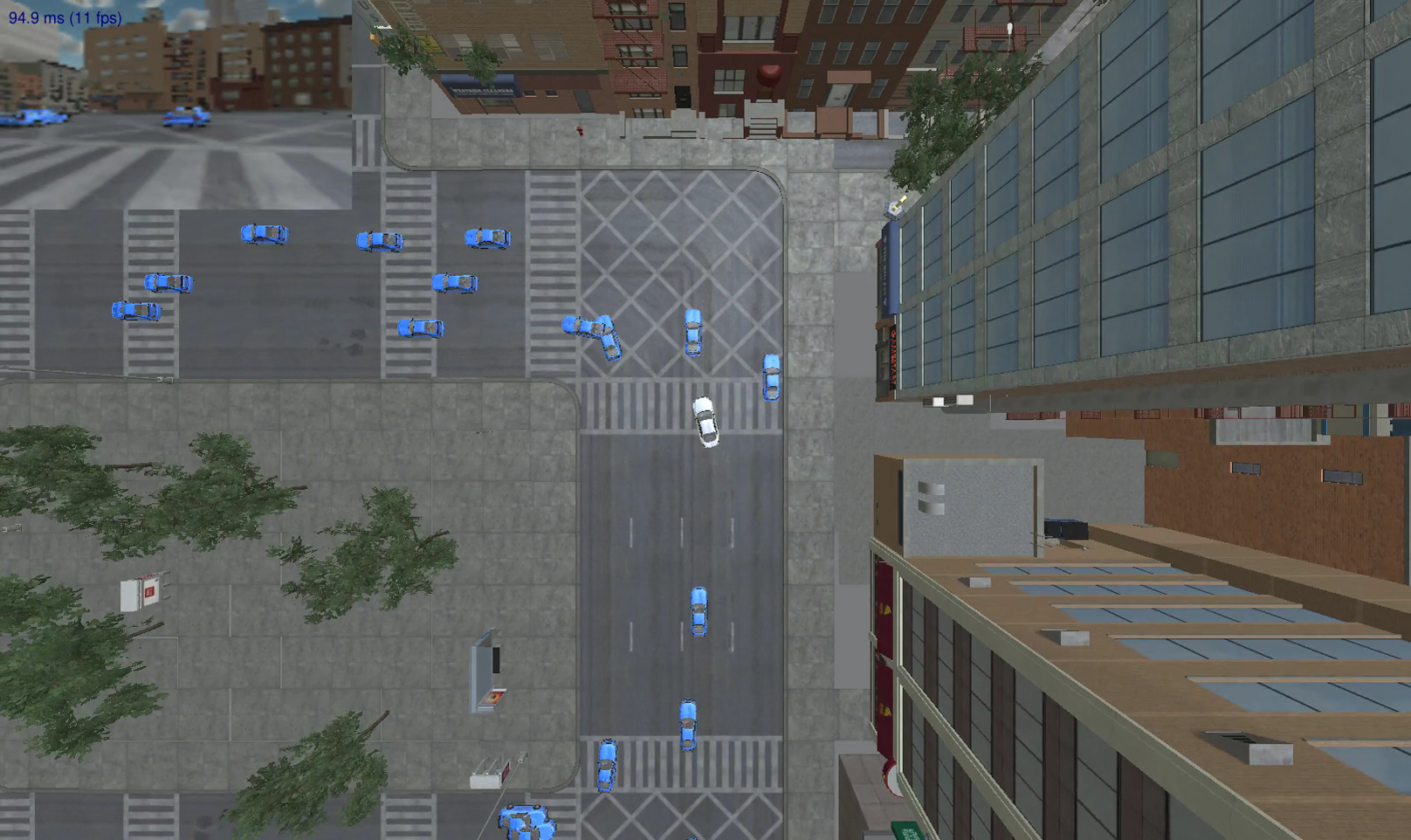}
    \caption{Illustration of multi-agent simulation in Unity}
    \label{fig:DRLtrafficCircle}
  \end{figure}

Currently, the standard approach for autonomous driving is to decouple the system into isolated sub-problems, typically supervised-learning-like object detection, visual odometry, etc and then having a post processing layer to combine all the results of the previous steps. There are two main issues with this approach: Firstly, the sub-problems which are solved may be more difficult than autonomous driving. For example, one might be solving object detection by semantic segmentation which is both challenging and unnecessary. Human drivers don't detect and classify all visible objects while driving, only the most relevant ones. Secondly, the isolated sub-problems may not combine coherently to achieve the goal of driving. In RL, this is explicitly handled by a reward signal corresponding to good driving which can model the interplay between driving (taking action) and planning (where to drive). As the reward is based on stable driving and not crashing typically, it is challenging to train an RL system with a real car because of the risks involved. Thus most of the current RL research is done using video game simulation engines like TORCS or Unity. Figure \ref{fig:DRLtrafficCircle} is a screen-shot of multi-agent simulation in Unity game engine which illustrates a difficult driving scenario where the white car tries to navigate in heavy traffic with sharp turns. This problem is relatively easier to model using RL. 

\subsection{MDP}

The model is developed under the Markov Decision Process (MDP) framework, which is a tuple of (S, A, $P_{sa}$, 
$\gamma$, R) where:
a)	Set of Environment states  (S)
b)	Set of Actions (A)
c)	Discount Factor ($\gamma$)
d)	Reward (R)
e)	State Transition Probabilities ($P_{sa}$)

We define a value-function that defines the value of being in a state S and following the policy $\pi$(s) till the end of the episode. The value-function is the expected sum of the discounted rewards as follows:
\begin{gather*}
V^\pi(s) = E[R(s_{0}) + \gamma R(s_{1}) + \gamma^2 R(s_{1}) + \dots \vert s_{0}=s, \pi(s)] 
\end{gather*}

The objective is to find the policy that maximizes the expectation of the accumulated rewards:
\begin{gather*}
J(\pi) = \max_{\pi} E[V^\pi(s)] \\
\pi^*(s) = \arg\max_{\pi} J(\pi)
\end{gather*}
 
The solution to such a problem lies in finding a policy $\pi$(s), which is formed of a set of actions A that maximizes the total rewards from a source to goal states S. There are popular algorithms to solve this problem for finite state-space like \textit{Value iteration} and \textit{Policy iteration} \cite{c24}.

\subsection{Q-learning}

Q-learning \cite{c30} is one of the commonly used algorithms to solve the MDP problem. The actions $a \in A$ are obtained for every state $s \in S$ based on an action-value function called $Q:S\times A \rightarrow \mathbb{R}$. The Q-learning algorithm is based on the Bellman equation:
\begin{gather*}
Q_{t+1}(s,a) \leftarrow Q_{t}(s,a) + \alpha[r +\gamma \arg\max_{a'} Q_{t}(s',a') - Q_{t}(s,a)]
\end{gather*}
The updates to the Q tables are done recursively by Temporal Difference (TD) incremental learning \cite{c24}. The algorithm starts from an initial state, and proceeds until the episode ends, i.e. a terminal state is reached. In every step, the agent is in the current state s, it takes an action following the policy $\pi$(s), and then observes the next state s’ together with the reward received from the environment r. The algorithm continues until convergence of the Q function or a certain number of episodes is reached.

\section{DEEP REINFORCEMENT LEARNING}

Depending on the problem domain, the space of possible actions may be discrete or continuous, a difference which has a profound effect on the choice of algorithms to be applied. In this section we will discuss two algorithms: one which operates on discrete actions (DQN) and one which operates on continuous actions (DDAC).  

\subsection{Deep Q Networks (DQN)} 

When the states are discrete, the Q-function can be easily formulated as a table. This formulation becomes harder when the number of states increases, and even impossible when the states are continuous. In such case, the Q-function is formulated as a parameterized function of the states, actions; $Q(s,a,w)$. The solution then lies in finding the best setting of the parameter $w$ . Using this formulation, it is possible to approximate the Q-function using a Deep Neural Network (DNN). The objective of this DNN shall be to minimize the Mean Square Error (MSE) of the Q-values as follows:

\begin{gather*}
l(w) = E[(r + \gamma \arg\max_{a'} Q_{t} (s',a',w) - Q_{t} (s,a,w))^2] \\
J(w) = \max_{w} l(w)
\end{gather*}
  
This objective function is differentiable end-to-end in with respect to its parameters, i.e. $\frac{\partial l(w)}{\partial w}$  exists. Thus, the optimization problem can be easily solved using Gradient-based methods (Stochastic Gradient Descent (SGD), Conjugate Gradients (CG),etc). The algorithm is called Deep Q-Networks (DQN) \cite{c17}\cite{c16}.
	
\subsection{Deep Deterministic Actor Critic (DDAC)} 

The DQN algorithm is suitable for continuous states cases, but action selection still requires the action values to be discrete. Several algorithms were suggested for continuous actions cases, where two functions are learned: 1) the actor; which provides the policy mapping from a state to action, and 2) the critic; which evaluates (criticizes) the value of the action taken in the given state. The critic function is the same as the Q-function. The algorithms to learn both functions follow the policy gradient methods \cite{c26}. Under the framework of deep learning, both functions can be learned through two neural networks; $Q(s,a,w)$  and $\pi(s,u)$ , since the whole objective is still differentiable w.r.t. the weights of the Q-function and the policy. Hence, the gradient of the Q-function (the critic) is obtained as in DQN: $\frac{\partial l(w)}{\partial w}$ , while the gradient of the policy function (the actor) is obtained using the chain rule as follows: 
\begin{gather*}
\frac{\partial J}{\partial u} = \frac{\partial Q}{\partial a} \vert_{a=\pi(s,u)} \frac{\partial \pi(s,u)}{\partial u}
\end{gather*}

 \subsection{Deep Recurrent Reinforcement Learning}

The Q-learning algorithms are based on the Markov assumption of the MDP. In situations where the full observability of the environment is not available, this assumption is not valid anymore. Partially observable MDP (POMDP) arises in different scenarios in autonomous driving, like the occlusion of objects during tracking, mapping and localization. POMDPs are tackled using information integration over time, where the true state of the environment is not directly revealed from single observation, but gradually form over multiple observations at different time stamps. The recurrent neural networks (RNN) present themselves as a natural framework to tackle POMDPs. In \cite{c19} RNN was successfully applied  for the task of end to end multi-object tracking. Moreover, LSTMs \cite{c8} are integrated to the DQNs to form the Deep Recurrent Q Networks (DRQN) in \cite{c7}. The application of DRQN in \cite{c7} to Atari games does not show the full power of the model, since the MDP assumption is usually enough for Atari games, hence the authors try a variant of the Pong game namely Flickering Pong, where they show the advantage of adding recurrence.

\subsection{Deep Attention Reinforcement Learning}

 In the DQN model, the spatial features are extracted via a CNN, which learns the features from data. These features may not be contributing equally to the final optimization objective. Similar to the recognition process in human beings, only a limited amount of information is needed to perform the recognition tasks, and not all the high dimensional sensory data. Attention models \cite{c32} are trying to follow the same concept, where only part of the CNN extracted features are used in the classification task. This part is learned in parallel to the original learning process. In \cite{c15}, a separate network called “glimpse network” is trained to deploy the kernel operation at certain parts of the image. The motion of the kernel is learned via the REINFORCE algorithm \cite{c31} to learn the best sequence of motions of the kernel over the input image. The result is a motion that resembles the reading process in human beings in case of feeding images of digits for example. The idea is exploited in \cite{c23} to improve the DQN and DRQN models, by including a filter after the CNN features extractor, resulting in the DARQN model. The DARQN was shown to attend and highlight parts of the Atari games that are of special importance. In the “Sequest” game, the agent was able to focus on the oxygen level of the submarine, while in “Breakout” game; the agent was able to track the ball position \cite{c31}.

\begin{figure}[htbp]
  \centering
  \includegraphics[scale=0.4]{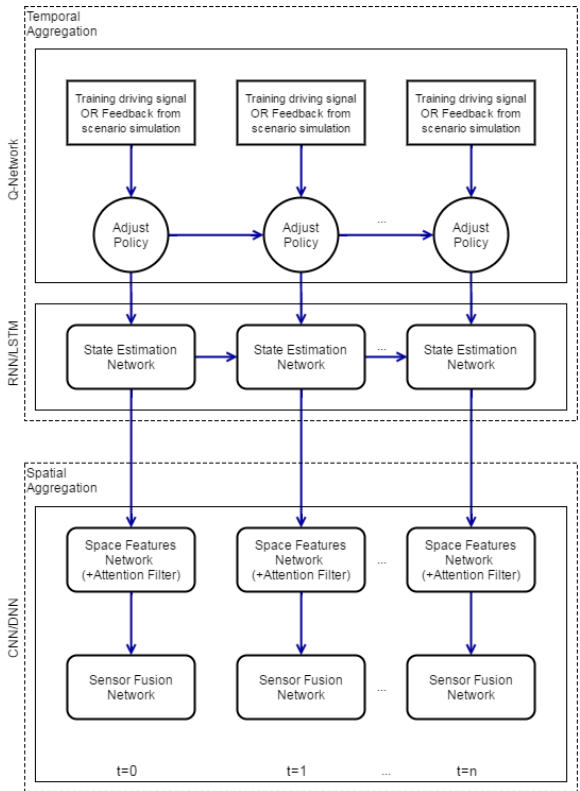}
  \caption{Deep Reinforcement Learning framework}
  \label{fig:DRLframework}
\end{figure}

\subsection{Apprenticeship learning}

The Reinforcement learning algorithms described so far follow the concept of “episodic” learning, or “learning from delayed rewards” \cite{c29}. In this setting, the rewards function is assumed to be known to the algorithm. However, in some cases, instead of having clear mapping of a state to reward function, we have a demonstrated expert behavior to the agent. The goal of the agent in this case is to decipher the intention of the expert, and decode the structure of the reward function. This is referred to as Inverse Reinforcement learning \cite{c1}. The reward function is encoded as a linear combination of “features”, functions that map the state according to some features of interest. For example, for the driving task, one feature function could detect “how far is the car from the lanes”. Another approach is described in \cite{c5}, where the demonstrated expert behavior is utilized in a supervised model, where the actions taken by the expert together with the states are considered as the training examples to a CNN.

\section{PROPOSED DRL FRAMEWORK FOR AUTONOMOUS DRIVING}

Following the models described for Deep Reinforcement Learning(DRL), we propose a pipelined framework for end-end training of a DNN for autonomous driving as illustrated in Figure \ref{fig:DRLframework}. The inputs are the states of the environment and their aggregations over time, and the output is the driving actions. We discuss each part in detail below with reference to a high-level block diagram in Figure \ref{fig:DRLblockDiagram}.

\subsection{Spatial aggregation}

The first step of the pipeline is the spatial aggregation network. This stage contains two networks: 1) Sensors fusion and 2) Spatial features, which are described below. The car state includes its position, orientation, velocity and acceleration. In case of autonomous driving, the surrounding environment state needs to be encoded as well. The environment state may include the objects, their location, orientation, motion vectors, dimensions,etc. Traditional autonomous driving algorithms would make use of these state vectors to plan the motion path in such an environment. The state of the surrounding objects is usually not directly observed, but rather deduced by the algorithm through a set of sensors readings (Camera, LIDAR, etc). Fusing such sensory information is critical for estimation of a robust state, which by itself is an active area of research. On the other hand, an end-end deep learning system would use a different encoding of the state. For example in \cite{c17}, the states are just the snapshots of the game, which include by default a lot of implicit information. Such information is not explicitly given to the algorithm, but rather deduced through the DNN (more specifically the CNN), in the form of “features”. 

\subsubsection{Sensor fusion}

This way of presenting the raw information to the DNN depends on its ability to extract relevant features, makes sensors fusion a natural task in the course of the learning process. In other words, it is possible to group all the sensors' information that encode the environment in a raw input vector, and present this to the DNN, leaving the fusion task to the weights adjustment process taking place inside the DNN using SGD for example. The gradients will take care of weighting each sensor features according to how relevant it is to minimizing the overall cost function discussed in the RL section. A CNN is best suited to this part of the pipeline. 

\subsubsection{Spatial features}

The fused sensors representation enables further processing of the data, focusing on the important part that is relevant to the driving task. A CNN would extract deeper representations, followed by an attention filter to direct the convolution kernels to the parts of the data that are of interest. The integration of attention models has a high potential for embedded systems integration. The attention models dramatically reduce the dimensionality of the data, in addition to the number of computations (convolutions) over the raw input data. 
To best utilize the value of attention models, it is recommended to use the idea of action and glimpse networks \cite{c15} rather than applying an attention filter as in \cite{c23}. The attention filter does not reduce the computations, where the convolution kernel is applied to all the data. On the other hand, the glimpse and action networks learn to only deploy the “gaze” of the kernel to certain parts of the data, thereby reducing the computations dramatically in inference time.

\tikzstyle{block} = 
[
draw = black!80, % black or black!80
very thick, % thick or very thick or semithick
%fill=blue!30, % blue!30 or green!50
top color=white,
bottom color=blue!30,
rectangle, 
rounded corners,
minimum height=6em, 
minimum width=10em, 
node distance= 4cm and 3cm,
align = center
]

\tikzstyle{data} = 
[
draw = black!80, % black or black!80
very thick, % thick or very thick or semithick
%fill=blue!30, % blue!30 or green!50
top color=white,
bottom color=blue!30,
trapezium,trapezium left angle=70,trapezium right angle=-70, 
%rounded corners,
minimum height=6em, 
minimum width=10em, 
node distance= 4cm and 3cm,
align = center
]

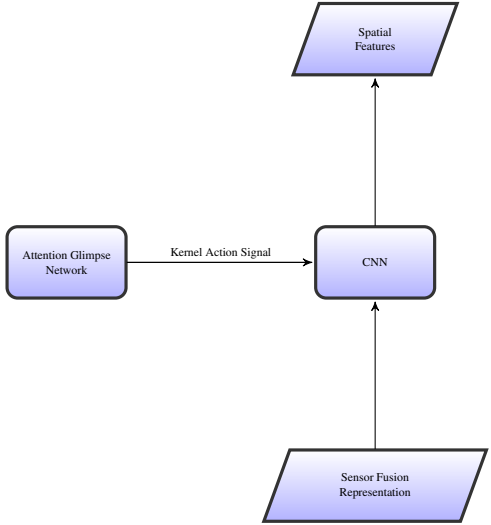
\begin{figure}
\centering
\beginpgfgraphicnamed{graphic-of-flat-world}
\begin{tikzpicture}
[
auto, 
>=stealth', 
shorten >=1pt ,
scale=0.5,
transform shape
]    

    %%  NODE PLACEMENTS - MAIN ROW
    \node [block    ](b1) {Attention Glimpse \\ Network};
    \node [block   ,right = 5 cm of b1 ](b2) {CNN};
    \draw (b1) -- (b2) node [midway, above] (TextNode) {Kernel Action Signal};

    %%  NODE PLACEMENTS - ROW ABOVE    
    \node [data   ,above = of b2 ](d1) {Spatial \\ Features };
    \node [data   ,below = of b2 ](d2) {Sensor Fusion \\ Representation };
    
    %% EDGES - MAIN ROW 
    \graph{ (b1) -> (b2) };

    %\draw [arrow] (d2)  -- (b2) {zz};
    %\draw (d2) edge [->] {aa} (b2);
    \draw [->] (d2)  -- (b2);
    \draw [->] (b2)  -- (d1);

\end{tikzpicture}
\endpgfgraphicnamed
\caption{Block diagram of modules in DRL framework}
\label{fig:DRLblockDiagram}
\end{figure}

\subsection{Recurrent temporal aggregation}

The environment state is not directly accessible to the agent or the autonomous car. Instead, sensor readings provide clues about the true state of the environment. To decode the true environment state, it is not sufficient to map single snapshot of sensors readings. However, information integration over time is mandatory, since the state is revealed as you move. A clear example is the famous robot localization problem. The most popular algorithm to solve such a problem is the Bayes filter algorithm, which depends on two probability density functions (pdf), usually known as the motion model and the sensor model. The Bayes filter is derived under the MDP assumption. When the state is discrete we have the Hidden Markov Model (HMM). Finally the Kalman filter is adding the linear state equations assumption, in addition to the Gaussian assumption over the two pdfs.

As discussed in the previous section, adding recurrence to the pipeline enables handling of POMDP situations. Such situations are often encountered in the driving scenarios. Traditional algorithms for tracking and localization, like Bayes filters, try to handle such situations through information integration over time. However, such algorithms are still derived under the framework of the MDP. Moreover, traditional signal processing approaches like Kalman filters require manual tuning of its model and parameters. Alternatively, Recurrent Neural Network (RNN) is able in theory to model long term dependencies on previous states, in addition to the current observation. In practice, due to limitations of the back propagation of gradients through time, RNN is not able to memorize long term dependencies, hence, the LSTM networks are proposed to solve such an issue \cite{c8}. LSTMs are able to control which information to keep from the previous state, in addition to which information to forget. Additionally, LSTM controls what to add to the hidden states from the new inputs.

It is worth noting that RNNs and LSTMs follow the same framework of Bayes filter, in terms of modeling the motion and observation models, where RNN is modeling the same pdfs using its hidden state weights and its input weights. In addition, all Bayes filters (and descendants like Kalman filters), RNN and LSTM have similar controls over what to include in the next hidden state from the old hidden state and from the new input. The general equation for Bayesian filtering and the comparison table of state and state transition functions are provided below.

\begin{gather*}
h_{t} = \eta p(x_{t} \vert h_{t})  \int p(h_{t} \vert h_{t-1}) h_{t-1} dh 
\end{gather*}

\setlength{\tabcolsep}{3pt}

\begin{tabular}{|c|c|c|c|}
\hline
Type   & $h_{t}$ & $p(x_{t}\vert h_{t})$ & $p(h_{t} \vert h_{t-1})$\\
\hline
RNN    & $tanh[(W_{hh}h_{t-1} + (W_{xh}x_{t})]$ & $W_{xh}$ & $W_{hh}$ \\
\hline
Kalman & $A_{t}h_{t-1} + K_{t}x_{t}$ & $K_{t}$  & $A_{t}$ \\
\hline 
LSTM   & $\bar{f_{t}}h_{t-1} + \bar{i_{t}}x_{t}$ & $\bar{i_{t}}$ & $\bar{f_{t}}$\\
\hline 
GRNN   & $\bar{f_{t}}h_{t-1} + \bar{i_{t}}x_{t}$ & $\bar{i_{t}}$ & $\bar{f_{t}}$\\
\hline 	
\end{tabular}
\captionof{table}{Comparison of state functions for temporal filters} 

\hfill 

\subsection{Planning}

\hspace{5mm} The final part of the end-end pipeline is the Reinforcement learning planning part. This network follows the same training procedure of the DQN, with a Q-network on the top having the same objective function as described in DQN section. In case of continuous actions, the DDAC network can be used as discussed in DDAC section. The same error is back propagated through the network structure to obtain the gradients at every network layer.

\section{RESULTS AND DISCUSSION}

\begin{figure}[htbp]
		\centering
		\includegraphics[width=1\columnwidth]{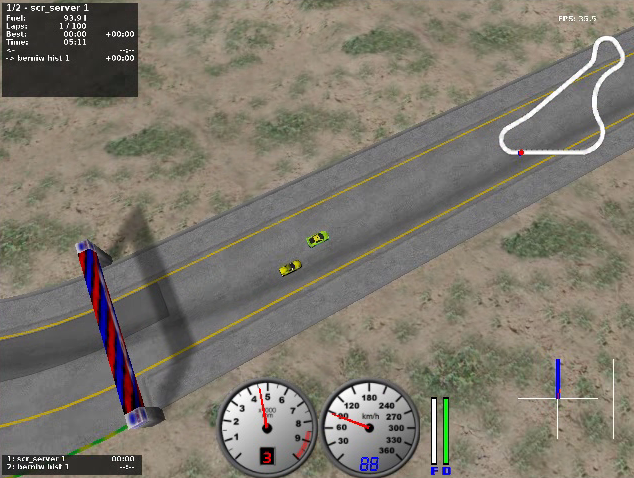}
		\caption{TORCS screen-shot of DRL based lane keeping}
		\label{fig:torcSimulation}
	\end{figure}

\hspace{5mm} We use lane keeping assist algorithm as a sample application to simulate and test the proposed framework on \textit{The Open-source Racing Car Simulator} (Torcs), screenshot shown in Figure \ref{fig:torcSimulation}. To be able to use simulated sensors, we use the \textit{Simulated Car Racing} (SCR) add-on, which gives access to the car controls, like steering, velocity, acceleration, brakes, and car states, like position, velocity, acceleration, fuel level, etc. In terms of simulation setup, the input to the network is the \textit{trackPos} sensor input, which provides the position of the track borders, in addition to the car speed in the x-position. The output are the steering, the gear, the acceleration and the brake values. The network is trained end-end following the same objective of the DQN. 

\hspace{5mm} In order to formulate the problem as a classification problem, the actions (steer, gear, brake, acceleration) are tiled and discretized. In another setup, the actions are taken as continuous values following the policy gradient method of the DDAC. The results show successful lane keeping function in both cases. However, in case of DQN, removing the replay memory trick (Q-learning) helps to have faster convergence and better performance. In addition, the tile coding of actions in case of DQN makes the steering actions more abrupt. On the other hand, using DDAC continuous policy helps smooth the actions and provides better performance.

\hspace{5mm} Apprenticeship learning with demonstrated behavior was also tested.  The experimental setup is similar to the DQN one described above. The demonstrated behavior is obtained from a proportional controller that is trying to keep the lane, while at the same time limiting the speed. The level of supervision is reduced gradually (the epsilon factor), so when epsilon is 0.5, it means 50\% of the time the actions are coming from the model, and 50\% from the p-controller. At epsilon=0, the model is completely in control. A regression neural network is used to fit the demo p-controller model. The result is a successful lane keeping behavior with speed limit. Youtube video links are provided in the appendix section for sample demonstration of training and lane keeping.

\section{CONCLUSION}

\hspace{5mm} In this paper, we provided a survey of the recent advances in the field of Deep Reinforcement Learning. We then followed by a proposal of a framework for an end-end Deep Reinforcement learning pipeline for autonomous driving which integrates RNNs to account for POMDP scenarios. We integrate attention models into the framework, making use of glimpse and action networks to direct the CNN kernels to the regions of the input that are relevant to the driving process, thereby improving the computational complexity. The framework was tested for lane keep assist algorithm and test videos with successful learning were provided. Our future work includes deploying the proposed framework in a simulated environment, where the sensors and actuators are artificially controlled with labeled ground truth. This can lead to potential extension of the framework to real driving scenarios.

%%%%%%%%%%%%%%%%%%%%%%%%%%%%%%%%%%%%%%%%%%%%%%%%%%%%%%%%%%%%%%%%%%%%%%%%%%%%%%%%
\section*{APPENDIX}

\hspace{5mm} Sample DRL training and demo sequences are provided as supplementary material for the review process. Please visit the following youtube links for \href{https://youtu.be/hktC8sGURJQ}{DRL training using DQN}, \href{https://youtu.be/OtuKpWew6UI}{DRL training using DDAC} and \href{https://youtu.be/RxIkdKGtzTE}{DRL driving demo using regression neural network}. \\

Entering URLs explicitly in case the hyperlinks are suppressed.
DRL training using DQN - https://youtu.be/hktC8sGURJQ \\
DRL training using DDAC - https://youtu.be/OtuKpWew6UI \\
DRL driving demo using regression neural network - https://youtu.be/RxIkdKGtzTE

\section*{ACKNOWLEDGMENT}

\hspace{5mm} The authors would like to thank their employer for the opportunity to work on fundamental research. Thanks to B Ravi Kiran (INRIA France) and Catherine Enright (Valeo) for reviewing the paper and providing feedback. Special thanks to a colleague Matthew Carrigan for detailed review, proof-reading and editing.

%%%%%%%%%%%%%%%%%%%%%%%%%%%%%%%%%%%%%%%%%%%%%%%%%%%%%%%%%%%%%%%%%%%%%%%%%%%%%%%%

%%%%%%%%%%%%%%%%%%%%%%%%%%%%%%%%%%
% Bibliography
%%%%%%%%%%%%%%%%%%%%%%%%%%%%%%%%%%

%%%%%%%%%%%%%%%%%%%%%%%%%%%%%%%%%%
% Biography
%%%%%%%%%%%%%%%%%%%%%%%%%%%%%%%%%%

\begin{biography}

Ahmad El Sallab is the Chief Engineer of Deep Learning in the Driving Active Safety unit, Valeo. Ahmad has 11 years of experience in Machine Learning and Deep Learning, where he acquired his M.Sc. and Ph.D. on 2009 and 2013 in the field. He has worked for reputable multi-national organization in the industry since 2005 like Intel and Valeo. He has 15 publications in Deep Learning in top IEEE and ACM journals and conferences, with applications in speech, NLP and Robotics. He is also a designated reviewer in a number of top conferences like EUSIPCO and ITSC.  \\

Mohammed Abdou is a software engineer and researcher in the Driving Active Safety unit, systems and functions team at Valeo Cairo. He is pursuing his master's degree in control systems with focus on Reinforcement Learning. \\

Etienne has been a research engineer in computer vision \& machine learning at Valeo Driving Assistance Research for 4 years where he applies Deep Learning extensively for perception and control. He focuses on embedded development, keeping in mind constraints from low-power devices. Etienne attended the French Engineering School of Polytech Orleans; his final specialisation year was spent at the University of Miami where he learned Machine Learning / Neural Networks from Miroslav Kubat.\\

Senthil Yogamani is a technical lead in computer vision department at Valeo for the past three years. He is currently focused on research and design of the overall computer vision algorithm architecture for next generation automated parking systems. He is also a member of the internal autonomous driving and deep learning working group. He has over 10 years of experience in computer vision and machine learning including 8 years of experience in industrial automotive systems. He is an author of 14 peer reviewed publications and technical reports. He is a recipient of best associate editor award at IEEE ITSC 2015 and best paper award at IEEE ITST 2012.

\end{biography}


\small
\begin{thebibliography}{99} 

\bibitem{c1} Abbeel, P., \& Ng, A. Y. (2004). Apprenticeship learning via inverse reinforcement learning. Proceedings of the twenty-first international conference on Machine learning, (p. 1).
% \bibitem{c2} Ba, J., Mnih, V., \& Kavukcuoglu, K. (2014). Multiple object recognition with visual attention. arXiv preprint arXiv:1412.7755 .
\bibitem{c3} Badrinarayanan, V., Kendall, A., \& Cipolla, R. (2015). SegNet: A Deep Convolutional Encoder-Decoder Architecture for Image Segmentation. arXiv preprint arXiv:1511.00561 .
% \bibitem{c4} Balduzzi, D., \& Ghifary, M. (2015). Compatible Value Gradients for Reinforcement Learning of Continuous Deep Policies. arXiv preprint arXiv:1509.03005 .
\bibitem{c5} Bojarski, M., Del Testa, D., Dworakowski, D., Firner, B., Flepp, B., Goyal, P., et al. (2016). End to End Learning for Self-Driving Cars. arXiv preprint arXiv:1604.07316 .
% \bibitem{c6} Hasselt, H. V. (2010). Double Q-learning. Advances in Neural Information Processing Systems, (pp. 2613-2621).
\bibitem{c7} Hausknecht, M., \& Stone, P. (2015). Deep recurrent q-learning for partially observable mdps. arXiv preprint arXiv:1507.06527 .
\bibitem{c8} Hochreiter, S., \& Schmidhuber, J. (1997). Long short-term memory. Neural computation , 9 (8), 1735-1780.
\bibitem{c9} Huval, B., Wang, T., Tandon, S., Kiske, J., Song, W., Pazhayampallil, J., et al. (2015). An Empirical Evaluation of Deep Learning on Highway Driving. arXiv preprint arXiv:1504.01716 .
%\bibitem{c10} Karavolos, D. (2013). Q-learning with heuristic exploration in Simulated Car Racing.
\bibitem{c11} Kendall, A., Badrinarayanan, V., \& Cipolla, R. (2015). Bayesian SegNet: Model Uncertainty in Deep Convolutional Encoder-Decoder Architectures for Scene Understanding. arXiv preprint arXiv:1511.02680.
\bibitem{c12} Krizhevsky, A., Sutskever, I., \& Hinton, G. E. (2012). Imagenet classification with deep convolutional neural networks. Advances in neural information processing systems, (pp. 1097-1105).
\bibitem{c13} LeCun, Y., Bengio, Y., \& Hinton, G. (2015). Deep learning. Nature , 521 (7553), 436-444.
%\bibitem{c14} Loiacono, D., Cardamone, L., \& Lanzi, P. L. (2013). Simulated car racing championship: Competition software manual. arXiv preprint arXiv:1304.1672 .
\bibitem{c15} Mnih, V., Heess, N., Graves, A., \& others. (2014). Recurrent models of visual attention. Advances in Neural Information Processing Systems, (pp. 2204-2212).
\bibitem{c16} Mnih, V., Kavukcuoglu, K., Silver, D., Graves, A., Antonoglou, I., Wierstra, D., et al. (2013). Playing atari with deep reinforcement learning. arXiv preprint arXiv:1312.5602 .
\bibitem{c17} Mnih, V., Kavukcuoglu, K., Silver, D., Rusu, A. A., Veness, J., Bellemare, M. G., et al. (2015). Human-level control through deep reinforcement learning. Nature , 518 (7540), 529-533.
%\bibitem{c18} Ng, A. Y., Russell, S. J., \& others. (2000). Algorithms for inverse reinforcement learning. Icml, (pp. 663-670).
\bibitem{c19} Ondruska, P., \& Posner, I. (2016). Deep tracking: Seeing beyond seeing using recurrent neural networks. arXiv preprint arXiv:1602.00991.
\bibitem{c20} Pinheiro, P. H., \& Collobert, R. (2013). Recurrent convolutional neural networks for scene parsing. arXiv preprint arXiv:1306.2795 .
\bibitem{c21} Sermanet, P., Eigen, D., Zhang, X., Mathieu, M., Fergus, R., \& LeCun, Y. (2013). Overfeat: Integrated recognition, localization and detection using convolutional networks. arXiv preprint arXiv:1312.6229 .
%\bibitem{c22} Silver, D., Lever, G., Heess, N., Degris, T., Wierstra, D., \& Riedmiller, M. (2014). Deterministic policy gradient algorithms. ICML. 
\bibitem{c23} Sorokin, I., Seleznev, A., Pavlov, M., Fedorov, A., \& Ignateva, A. (2015). Deep Attention Recurrent Q-Network. arXiv preprint arXiv:1512.01693 .
\bibitem{c24} Sutton, R. S. (1988). Learning to predict by the methods of temporal differences. Machine learning , 3 (1), 9-44.
\bibitem{c25} Sutton, R. S., \& Barto, A. G. (2016). Reinforcement learning: An introduction. Online Draft.
\bibitem{c26} Sutton, R. S., McAllester, D. A., Singh, S. P., Mansour, Y., \& others. (1999). Policy Gradient Methods for Reinforcement Learning with Function Approximation. NIPS, 99, pp. 1057-1063.
% \bibitem{c27} Uijlings, J. R., van de Sande, K. E., Gevers, T., \& Smeulders, A. W. (2013). Selective search for object recognition. International journal of computer vision , 104 (2), 154-171.
% \bibitem{c28} Van Hasselt, H., Guez, A., \& Silver, D. (2015). Deep reinforcement learning with double Q-learning. arXiv preprint arXiv:1509.06461 .
\bibitem{c29} Watkins, C. J. (1989). Learning from delayed rewards. Ph.D. dissertation, University of Cambridge England.
\bibitem{c30} Watkins, C. J., \& Dayan, P. (1992). Q-learning. Machine learning , 8 (3-4), 279-292.
\bibitem{c31} Williams, R. J. (1992). Simple statistical gradient-following algorithms for connectionist reinforcement learning. Machine learning , 8 (3-4), 229-256.
\bibitem{c32} Xu, K., Ba, J., Kiros, R., Courville, A., Salakhutdinov, R., Zemel, R., et al. (2015). Show, attend and tell: Neural image caption generation with visual attention. arXiv preprint arXiv:1502.03044 .
%\bibitem{c33} Zambrano, D., Roelfsema, P. R., \& Bohte, S. M. (2015). Continuous-time on-policy neural reinforcement learning of working memory tasks. Neural Networks (IJCNN), 2015 International Joint Conference on, (pp. 1-8).

\end{thebibliography}
\end{document}